%% file: main.tex
\documentclass[conference,a4paper]{IEEEtran}
\IEEEoverridecommandlockouts

\usepackage{cite}
\usepackage{amsmath,amssymb,amsfonts}
\usepackage{algorithmic}
\usepackage{graphicx}
\usepackage{textcomp}
\usepackage{xcolor}
\usepackage{booktabs}
\usepackage{multirow}
\usepackage{dblfloatfix}
\usepackage{subcaption}
\usepackage{url}
\def\BibTeX{{\rm B\kern-.05em{\sc i\kern-.025em b}\kern-.08em
    T\kern-.1667em\lower.7ex\hbox{E}\kern-.125emX}}
\begin{document}

\title{Temporal Modelling for Burn Scars on Sentinel-3
\thanks{This work has been carried out in the context of the following projects: UNICORN (G.A. 101180172), REHUBS (G.A. 101214051) and Space It Up funded by the Italian Space Agency and the Ministry of University and Research - Contract No. 2024-5-E.0 - CUP No. I53D24000060005.}
}

\author{
\IEEEauthorblockN{
Luca Barco\IEEEauthorrefmark{1}\IEEEauthorrefmark{2},
Edoardo Arnaudo\IEEEauthorrefmark{2},
Andrea Bragagnolo\IEEEauthorrefmark{2},
Claudio Rossi\IEEEauthorrefmark{2},
Paolo Garza\IEEEauthorrefmark{1}
}
\IEEEauthorblockA{\IEEEauthorrefmark{1}\textit{Politecnico di Torino, Torino, Italy} \\
\{name.surname\}@polito.it}
\IEEEauthorblockA{\IEEEauthorrefmark{2}\textit{Fondazione LINKS, Torino, Italy} \\
\{name.surname\}@linksfoundation.com}
}
\maketitle
\begin{abstract}
Rapid and accurate burn scar delineation from satellite imagery is essential for post-fire damage assessment. Sentinel-3 OLCI, with daily revisit and 21 spectral bands, suits rapid mapping, yet most pipelines treat acquisitions independently, leaving the pre/post-fire change signal unexploited. We present a dataset of 246 wildfire activations (2016-2025) from the Copernicus Emergency Management Service, with Sentinel-3 OLCI temporally paired acquisitions. We benchmark spatial and temporal (ConvLSTM-augmented) variants of three backbones (U-Net, SegFormer, ConvNeXt-UPerNet) under two input modes and spectral configurations. Temporal modeling improves segmentation only when pre-fire frames are included, and a 5-band subset matches the full 21-band OLCI configuration.
\end{abstract}


\input{sections/01_intro}
\input{sections/02_related}
\input{sections/03_dataset}
\input{sections/04_methods}
\input{sections/05_exps}
\input{sections/06_conclusions}
\bibliography{refs.bib}

\end{document}

%% file: sections/01_intro.tex
\section{Introduction}
\label{sec:intro}

Wildfires pose an escalating threat to ecosystems, infrastructure, and human life, with burned area and fire frequency increasing markedly under climate change. Rapid burn scar delineation is critical for damage assessment and emergency response, and Sentinel-3 OLCI, with daily revisit, 300\,m resolution and 21 visible-to-near-infrared bands, is well suited to it. Despite this temporal richness, most existing burn scar detection pipelines treat imagery as independent snapshots: a single post-fire scene or a pre/post stacked composite is fed to a segmentation model, discarding the multi-temporal progression that frequent-revisit satellites inherently provide. This snapshot paradigm leaves a natural change signal unexploited, namely the spectral contrast between pre-fire vegetation and post-fire charred surface, which can be obscured in any single acquisition by smoke, cloud cover, or atmospheric variability.

This paper investigates whether explicit temporal modeling improves burn scar segmentation compared to spatial-only baselines when using Sentinel-3 OLCI imagery. Our contributions are: (i) We release \textbf{TMB-S3} (\textbf{T}emporal \textbf{M}odelling for \textbf{B}urn scars on \textbf{S}entinel-\textbf{3})\footnote{Code available at \url{https://github.com/links-ads/tmb-s3}. Dataset available at \url{https://huggingface.co/datasets/links-ads/tmb-s3} .} a large-scale burned-area dataset of 246 European wildfire activations (2016-2025) derived from the CEMS archive, comprising temporally paired Sentinel-3 OLCI acquisitions with geographically stratified splits; 
(ii) We augment U-Net, SegFormer, and ConvNeXt-UPerNet with a ConvLSTM temporal encoder and show via ablation that recurrent modeling helps only when a pre-fire reference is included, outperforming a bi-temporal early-fusion baseline, and that a five-band surface subset matches the full 21-band configuration under that conditioning.

%% file: sections/02_related.tex
\section{Related Works}
\label{sec:related}

\textbf{Deep learning for burned-area segmentation.}
Burned-area mapping has traditionally relied on spectral indices, chiefly the Normalized Burn Ratio and its bi-temporal difference (dNBR)~\cite{key2006nbr}, which underpin operational global products such as MODIS MCD64A1~\cite{giglio2018mcd64} and FireCCI51~\cite{lizundia2020firecci}. Deep learning has since recast the task as semantic segmentation. The dominant setting is mono-temporal, typically training a U-Net on single post-fire Sentinel-2 scenes ~\cite{knopp2020unet,seydi2022burntnet,farasin2020double}. A second family stacks a pre- and post-fire pair as a multi-channel input or frames detection as bi-temporal change detection~\cite{zhang2021synergy, sui2024biaunet}. In both cases imagery is treated as one or two static snapshots, and the multi-temporal progression offered by frequent-revisit sensors is left unexploited.

\textbf{Sensors and datasets.}
This snapshot bias is partly driven by sensor choice. High-resolution optical missions (Sentinel-2, 10-20\,m) dominate burned-area mapping, but their multi-day revisit limits temporal sampling; daily-revisit sensors such as MODIS and VIIRS trade spatial detail for temporal density, supporting learned pipelines for mapping and dating~\cite{pinto2020banet} and coarse-to-fine fusion~\cite{pinto2021practical}. Sentinel-3 occupies this regime and its SLSTR channels support operational active-fire and FRP retrieval~\cite{xu2023slstr}, yet OLCI remains, to our knowledge, largely unexploited for learned burned-area segmentation. Public benchmarks reflect the same bias: datasets such as FLOGA~\cite{sdraka2024floga}, CaBuAr~\cite{cambrin2024cabuar}, and Copernicus EMS-derived collections ~\cite{colomba2022dataset,arnaudo2023robust} are Sentinel-2/Sentinel-1 single- or paired-snapshot sets. No large, temporally-paired Sentinel-3 OLCI burned-area dataset exists: a gap our dataset addresses.

\textbf{Temporal modeling in Earth observation.}
Temporal deep learning is instead mature in adjacent EO tasks. Recurrent convolutional encoders model phenological dynamics in Sentinel-2 time series for land-cover classification~\cite{russwurm2018recurrent}, while temporal self-attention models~\cite{garnot2020psetae,garnot2021utae} set the state of the art for crop mapping, and bi-temporal transformers advance generic change detection~\cite{chen2021bit}. For wildfires, temporal models have mainly addressed spread forecasting~\cite{gerard2023wildfirespreadts}; whether recurrent encoders benefit burned-area segmentation, and whether they need a pre-fire reference, remains unstudied. 

%% file: sections/03_dataset.tex
\section{Dataset}
\label{sec:dataset}

Our burned-area segmentation dataset is built from 246 European wildfire activations in the Copernicus Emergency Management Service (CEMS) archive, from 2016 to 2025. For each activation, we collect fire extent polygons, sensor image footprints, validity masks together with curated event metadata (activation time and pre/post images acquisition dates to produce fire polygons) and land cover information from ESA WorldCover \cite{zanaga2022worldcover}.

Because a single event can include multiple partially overlapping AOIs, we merge spatially consistent AOIs and intersect them with the corresponding sensor footprint so that only actually observed areas are retained. We then tile each geometry in its native UTM projection into 25 km $\times$ 25 km bounding boxes (bboxes). Bboxes generated from different tiles are deduplicated when they share the same event time and overlap by more than 30\%; in those cases, we keep the bbox with lower expected projection distortion. For every bbox, we produce two annotation layers at 10 m resolution: a binary fire mask and a binary validity mask. Because Sentinel-3 OLCI imagery is at 300 m, these 10 m masks are resampled to the OLCI grid by logical union: a 300 m pixel is labeled burned if any of its underlying 10 m pixels is burned. This recall-favoring rule inflates the burned class at coarse resolution: burned pixels account for 21.0\% of valid pixels under the union rule, against 15.7\% when measured as true fractional burned area, i.e., a relative inflation of 34\%, mostly due to a one-pixel dilation of the fire boundary. The effect depends strongly on fire size: burned area is inflated by about $3\times$ for fires smaller than 10 OLCI pixels ($<$0.9\,km$^2$), by $1.76\times$ for 10-100 pixels, and by $1.25\times$ above 1000 pixels. A majority rule (burned fraction $\geq 0.5$) would preserve the total area ($1.02\times$) but remove most of the smallest fires, whose partially burned pixels still carry a detectable spectral change. The same rule is applied to training and test labels, so the comparison between models is unaffected, whereas absolute F1 and IoU are measured against a slightly dilated perimeter. 

Pre-event and post-event imagery are then retrieved with metadata-driven temporal windows of 30 days before and 90 days after each event, extended when no valid acquisition is found. The extension is frequently needed for the pre-fire frame, whose median lead time is 31 days (IQR 8-161). This asymmetry reflects the assumption that pre-event vegetation is comparatively stable, so that a single cloud-free acquisition can serve as baseline; for the longest lead times, however, seasonal or phenological changes may also be present in the pre/post contrast. The longer post-event window compensates for smoke and fire-induced clouds, which can obscure the scar for weeks. The dataset includes Sentinel-3 OLCI radiance products that are converted to top-of-atmosphere reflectance; cloud masks are derived from native quality flags, and acquisitions with more than 20\% invalid pixels are discarded. This yields temporally paired, quality-controlled observations for each bounding box. Every bbox has exactly one pre-fire acquisition and at least one post-fire acquisition. The first post-fire frame is acquired a median of 0.6 days after the event date, and the post-fire sequence ends at the first clear acquisition after the CEMS delineation timestamp (median lag $+1.5$ days; 63\% of bboxes within 3 days and 90\% within 11 days), so that the final frame depicts the consolidated burn scar used as ground truth. Only 18 bboxes (2.3\%) exceed 30 days; in these cases, residual fire growth or vegetation regrowth after the delineation are a possible source of label noise.

\begin{figure}
    \centering
    \includegraphics[width=\linewidth]{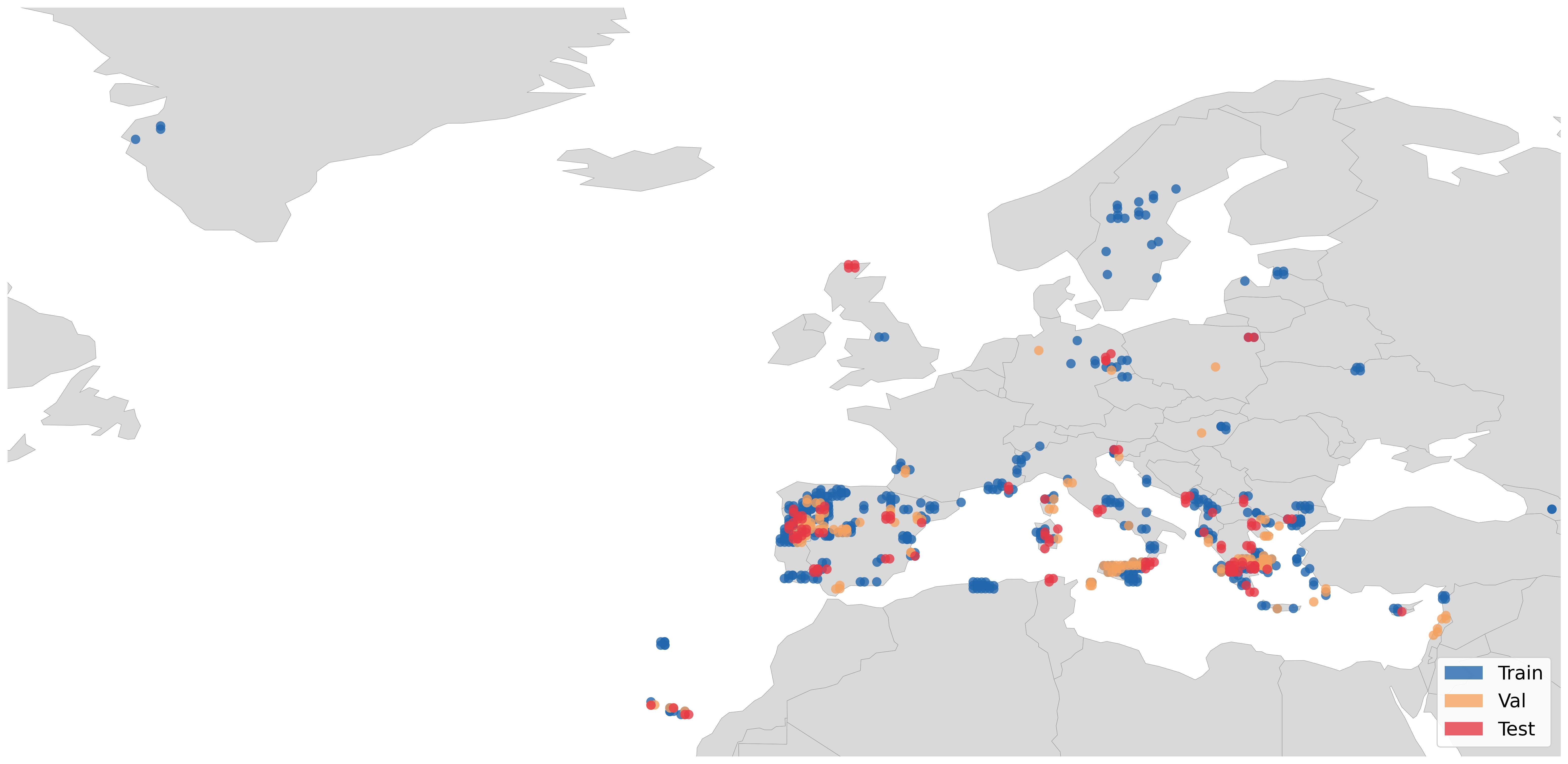}
    \caption{Geographic distribution of the dataset events: 146 train, 50 validation, and 50 test events.}
    \label{fig:dataset_map}
\end{figure}
All modalities and labels are stored per bbox with harmonized geospatial metadata to support consistent training and evaluation. Split assignment is performed at the event level, so that all bboxes of an event belong to the same split. Events are clustered geographically with DBSCAN ($\varepsilon{=}150$\,km), and the combination of cluster and event-size bin is used as stratification label to partition events into train/validation/test at 60/20/20. This stratification ensures that each split covers all regions and fire sizes; as a consequence, geographically close events from the same cluster may fall in different splits. Finally, the dataset comprises 146 events (517 bboxes) for train, 50 events (135 bboxes) for validation and 50 events (118 bboxes) for test. Each bbox is $\approx$25\,km $\times$ 25\,km, i.e., 83-90 pixels per side at 300\,m, events span a median of 2 bboxes (max 48), and sequences contain a median of $T{=}9$ acquisitions (IQR 6-12, range 2-36), i.e., one pre-fire and a median of 8 post-fire frames; median $T$ is 9, 10 and 8 for train, validation and test, respectively. Valid pixels cover 26.6\% of the bbox area overall, and burned pixels account for 21.4\%, 17.5\% and 24.2\% of valid pixels in the three splits (15.9\%, 13.1\% and 18.2\% as fractional area). 

%% file: sections/04_methods.tex
\section{Methodology}
\label{sec:methods}

\subsection{Problem Formulation}

We frame burned area delineation as a binary segmentation task. A single satellite acquisition at time $t$ is represented as $\mathbf{x}_t \in \mathbb{R}^{C \times H \times W}$, where $H \times W$ is the spatial extent of a patch and $C = B + 1$ is the number of input channels: $B$ Sentinel-3 OLCI bands ($B{=}21$ for the full-band and $B{=}5$ for the band-subset configuration) plus one static ESA WorldCover~\cite{zanaga2022worldcover} land-cover channel, i.e., $C{=}22$ or $C{=}6$. Each wildfire event is associated with a temporally ordered sequence of $T$ acquisitions $\mathbf{X} = (\mathbf{x}_1, \ldots, \mathbf{x}_T)$; the sequence length $T$ varies across events and includes exactly one \emph{pre-fire} acquisition, taken before the event began, followed by one or more \emph{post-fire} acquisitions, taken from the fire onset onwards (possibly while the fire is still active). The model produces pixel-wise burn probabilities $\hat{\mathbf{y}} \in [0,1]^{H \times W}$, supervised by binary burned area masks $\mathbf{y} \in \{0,1\}^{H \times W}$.

We compare two processing paradigms. In the \textbf{2D (spatial)} mode, a spatial segmentation network receives a single input tensor, built from at most two acquisitions (Section~\ref{sec:modes}), and predicts the burned area at the final acquisition. In the \textbf{3D (temporal)} mode, frames are processed in temporal order by a recurrent encoder that accumulates a hidden state before forwarding per-frame representations to the segmentation head, thereby preserving the temporal structure of the sequence.
\subsection{Segmentation Architectures}

We evaluate three spatial segmentation backbones. Used directly on a single input they serve as 2D baselines; wrapped by the temporal encoder described in Section~\ref{sec:temporal}, they form the corresponding temporal variants.

\textbf{U-Net.} A lightweight encoder-decoder network trained from scratch. The encoder comprises three downsampling stages, each applying two successive $3 \times 3$ convolutions with batch normalisation and ReLU activation, followed by max-pooling. Channel widths progress from 32 to 64 to 128, with a bottleneck of 256 channels.

\textbf{SegFormer.} A lightweight SegFormer~\cite{xie2021segformer} trained from scratch. Three encoder stages apply overlapping patch embeddings with stride 2, halving the spatial resolution at each stage. Each stage employs multi-head self-attention with spatial reduction (reduction ratios 4, 2, and 1 respectively) and feed-forward layers with $3 \times 3$ depthwise convolution; stage depths are $(2, 2, 2)$ and embedding dimensions $(32, 64, 160)$. 

\textbf{ConvNeXt-UPerNet.} A ConvNeXt-Tiny encoder~\cite{liu2022convnext} pretrained on ImageNet, combined with a UPerNet~\cite{10.xiao2018upernet} segmentation head. The first convolutional layer is adapted to the number of input channels $C_{\mathrm{in}}$ of each configuration by cyclically repeating the pretrained RGB weights across all input channels and rescaling by $3/C_{\mathrm{in}}$ to preserve activation magnitude; all remaining weights are retained from the ImageNet checkpoint.

\subsection{ConvLSTM Temporal Encoder}
\label{sec:temporal}

For temporal processing, each backbone is extended with a ConvLSTM encoder~\cite{convlstm} that propagates spatial context across the input sequence. A single ConvLSTM cell with $d_h = 64$ hidden channels and a $3 \times 3$ convolutional kernel is unrolled over the $T$ frames of the sequence. The purpose of the temporal processing is to accumulate spectral evidence across acquisitions and follow the evolution of the fire: post-fire frames captured on different dates may differ in burned area extension, cloud cover or viewing angle, and the pre-fire frame provides a reference of the landscape before the event.
No explicit indicator of each frame's role is given to the model: since the pre-fire acquisition, when included, is always the first element of the sequence, the recurrence can identify it from its position. In preliminary experiments with U-Net, appending a binary pre/post indicator as an additional ConvLSTM input channel changed the pre- and post-fire F1 by less than 0.8 points, within one standard deviation, so we omit it. For example, for an event with one pre-fire acquisition followed by three post-fire acquisitions ($T=4$), the cell processes the sequence $(\mathbf{x}_1^{\mathrm{pre}},\, \mathbf{x}_1^{\mathrm{post}},\, \mathbf{x}_2^{\mathrm{post}},\, \mathbf{x}_3^{\mathrm{post}})$, so that the hidden state encodes the pre-fire baseline before integrating the post-fire evidence.
At each step $t$, the hidden state $\mathbf{h}_t \in \mathbb{R}^{d_h \times H \times W}$ is concatenated with the original frame $\mathbf{x}_t$ along the channel axis, yielding a $(C + d_h)$-channel representation forwarded to the 2D backbone. All frames in the batch are processed by the backbone in a single parallel pass. Loss and evaluation metrics are computed on the \emph{final} post-fire acquisition only, which is always acquired after the CEMS delineation timestamp and therefore depicts the consolidated burn perimeter used as ground truth (Section~\ref{sec:dataset}). All preceding frames (i.e., the pre-fire frame, when included, and any intermediate post-fire frames) contribute only through the forward pass and the accumulation of the recurrent hidden state, receiving no direct gradient; their per-timestep predictions are used solely for qualitative visualisation.

\textbf{Variable-length batching.} For the temporal models, frames are sorted by acquisition time, with the pre-fire frame first. Within a training batch, sequences are right-padded with zero frames to the longest sequence, and a boolean validity mask records the real time steps. The loss is computed only at the last valid step of each sequence. Since the ConvLSTM is unidirectional and padding follows the real frames, padded steps cannot influence the hidden state at any valid step. Validation and test are run with batch size 1, hence without padding.

\subsection{Temporal Conditioning Modes}
\label{sec:modes}

We evaluate two input configurations that differ in the temporal context available to the model:

\textbf{Post-fire only.} The input sequence $\mathbf{X} = (\mathbf{x}_1, \ldots, \mathbf{x}_T)$ contains only acquisitions taken from the fire onset onwards, with the pre-fire reference removed. Since every bbox has a pre-fire acquisition, this mode is an ablation on the same sequences rather than a separate subset of events. The model detects burned areas from spectral signatures in these images alone.

\textbf{Pre- and post-fire.} The input sequence contains one pre-fire acquisition ($\mathbf{x}_1$), taken before the event began, prepended to the $T-1$ acquisitions from the fire onset onwards. This enables the model to exploit the spectral change signal between the pre-event reference and subsequent imagery as an additional cue for burned area delineation.

The temporal (3D) models process the full sequence $\mathbf{X}$ recurrently. The spatial (2D) models take a single input and predict the burned area at the final acquisition $\mathbf{x}_T$: in the post-fire only setting the input is $\mathbf{x}_T$; in the pre- and post-fire setting, the pre-fire reference $\mathbf{x}_1$ and $\mathbf{x}_T$ are concatenated along the channel dimension (bi-temporal early fusion), with the static land-cover channel included once, i.e., $2B{+}1$ input channels ($43$ or $11$). The 2D models therefore do not use the intermediate acquisitions $\mathbf{x}_2, \ldots, \mathbf{x}_{T-1}$, and the gap between the 2D and 3D pre- and post-fire results reflects the benefit of modelling the full post-fire sequence recurrently.

\subsection{Training Objective}

All models are trained with pixel-wise masked binary cross-entropy $
    \mathcal{L} = \mathcal{L}_{\mathrm{BCE}}(\hat{\mathbf{y}},\, \mathbf{y};\, w),
$
where $\mathbf{y}$ is the binary fire mask, $\hat{\mathbf{y}}$ the model prediction, and $w \in \{0, 1\}$ a binary pixel mask that excludes invalid observations from the loss. The loss applies no class re-weighting; the fire/background imbalance is instead partially mitigated by the burned-pixel-centred cropping described in Section~\ref{sec:exps}.

%% file: sections/05_exps.tex
\section{Experiments}
\label{sec:exps}
\subsection{Experimental Setup}
\input{tables/all}
\begin{figure*}[!t]
    \centering
    \begin{subfigure}[t]{0.55\linewidth}
        \centering
        \includegraphics[width=\linewidth]{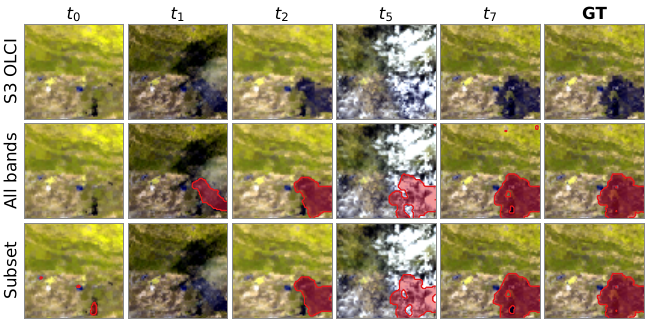}
        \caption{EMSR457 - AOI 1.}
        \label{fig:emsr457}
    \end{subfigure}
    \hfill
    \begin{subfigure}[t]{0.35\linewidth}
        \centering
        \includegraphics[width=\linewidth]{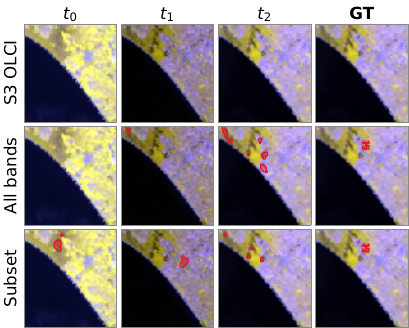}
        \caption{EMSR454 - AOI 4.}
        \label{fig:emsr454}
    \end{subfigure}
    \caption{Qualitative results. S3 OLCI input timeseries with related predictions using full-band and band-subset configurations. In each timestep column, red pixels overlaid on the S3 input indicate predicted fire pixels; the final GT column shows the ground-truth burned area the model is expected to predict. (a) Successful delineation of the growing burned area. (b) Failure case: small burned area with limited valid pixel coverage. For EMSR457, five of the eight acquisitions are shown; $t_0$ is the pre-fire acquisition.}
    \label{fig:qualitatives}
\end{figure*}
\textbf{Training protocol.} Training uses random $64 \times 64$ crops ($\approx$19.2\,km), one per bbox per epoch: 50\% are centred on a randomly selected burned pixel and 50\% placed uniformly at random, so that crops with little or no burned area are also seen; overall, 79\% of the valid pixels are unburned. The benchmark comprises the three backbones in 2D and 3D form, two conditioning modes (Section~\ref{sec:modes}), two spectral configurations (all $B{=}21$ OLCI bands vs the $B{=}5$ blue to near-infrared subset Oa04, Oa06, Oa08, Oa17, Oa21), and three random seeds (17, 42, 127), for a total of 72 training runs. All models are optimised with AdamW (learning rate $10^{-3}$, weight decay $10^{-4}$), cosine annealing over 100 epochs, batch size 8 and gradient clipping at 1.0; the checkpoint with the best validation F1 is retained, validation being used only for this purpose.

\textbf{Evaluation.} F1 and IoU on the fire class are computed on the final post-fire acquisition of each test bbox, over valid pixels within the CEMS validity mask, with predictions on permanent water (ESA WorldCover) set to unburned. Pixels are classified as burned when the predicted probability is at least 0.5, a threshold fixed a priori and not tuned. Inference runs on the full bbox with overlapping $64 \times 64$ sliding windows (stride 32, spline blending), so false alarms over the unburned surroundings of each fire are counted; false alarms on fire-free scenes and outside CEMS-mapped areas are not assessed. Metrics are computed per bbox and averaged, unweighted, over the 118 test bboxes. As events contribute 1 to 7 bboxes, we also computed per-event scores (averaged within each event, then over the 50 events): per-event F1 differs from per-bbox F1 by at most 1 point, and all findings below hold. Results are mean $\pm$ sample standard deviation over three seeds.

\subsection{Results and Discussion}
Table~\ref{tab:all} reports F1 and IoU on the test set for all configurations.
\paragraph{Temporal encoding requires a pre-fire reference to be effective}
The central finding of this work is that temporal modelling yields consistent and substantial gains only when pre-fire acquisitions are included in the input sequence. For each of the three architectures and both spectral configurations, the ConvLSTM temporal encoder with pre- and post-fire frames outperforms all other settings of the same architecture, achieving the best overall F1 of $67.64\pm0.91$ (SegFormer, full-band) and $68.24$ (band-subset), reached by both U-Net ($\pm1.72$) and SegFormer ($\pm1.19$). The corresponding absolute improvements over the strongest spatial baseline, the bi-temporal SegFormer ($65.50$ and $65.73$), are $+2.14$ and $+2.51$ F1 points, respectively ($+3.68$ and $+4.60$ over the strongest post-fire-only spatial baseline, SegFormer with $63.96$ and $63.64$). By contrast, temporal encoding applied to post-fire frames alone provides negligible or no improvement: the largest gain is $+2.96$ F1 (ConvNeXt, band-subset), and all other differences from the spatial counterpart, positive or negative, are within one standard deviation. A plausible explanation is that post-fire frames share the same burned-area signature, so their ordering creates no change signal and the hidden state mostly absorbs frame-to-frame variability from atmosphere, clouds and sensor noise; with a pre-fire frame, it can instead propagate the contrast between healthy vegetation and charred surface, a far stronger cue.
The gains cannot be attributed to added capacity alone. The ConvLSTM adds 0.22\,M trainable parameters to U-Net (1.93\,M $\rightarrow$ 2.15\,M) and SegFormer (1.03\,M $\rightarrow$ 1.25\,M), and 0.30\,M to ConvNeXt-UPerNet (37.07\,M $\rightarrow$ 37.37\,M, $<$1\%). For U-Net and SegFormer, the same encoder, with the same parameter count, yields no significant gain in post-fire-only mode (e.g., $62.63$ vs $62.20$ and $63.91$ vs $63.96$, full-band), and the 37\,M-parameter ConvNeXt is outperformed by the much smaller U-Net and SegFormer in every matched setting.

\paragraph{The pre-fire reference helps spatial models, the full sequence helps further}
The bi-temporal early-fusion baselines confirm that the pre-fire reference is informative even without recurrence: concatenating the pre-fire and the final post-fire acquisition improves over the single post-fire frame for every architecture and spectral configuration, by $+1.27$ to $+2.85$ F1. Modelling the full sequence with the ConvLSTM brings a further, consistent gain over the bi-temporal baselines, from $+1.09$ (ConvNeXt, band-subset, within one standard deviation) to $+3.60$ F1 (U-Net, band-subset). Since the two pre+post configurations share the same pre-fire reference and the same final acquisition, this additional gain is likely attributable to the intermediate post-fire acquisitions, which the recurrent encoder can integrate to follow the evolution of the burned area and to compensate for partially occluded frames; however, the two configurations also differ in fusion mechanism (channel concatenation vs recurrence), so the two effects cannot be fully separated.

\paragraph{Band-subset matches full-band for temporal pre+post models}
Contrary to the usual assumption that more spectral information is better, the band-subset configuration under temporal pre+post conditioning matches the full-band configuration for all three architectures: $68.24$ vs $66.65$ F1 for U-Net ($+1.59$), $68.24$ vs $67.64$ for SegFormer ($+0.60$), and $59.80$ vs $59.98$ for ConvNeXt ($-0.18$), all differences being within one standard deviation. The same holds for the bi-temporal 2D baselines ($64.64$ vs $64.63$ for U-Net, $65.73$ vs $65.50$ for SegFormer, $58.71$ vs $57.26$ for ConvNeXt).
We hypothesise that this reflects spectral redundancy: the subset retains a broadband blue-green-red-NIR sampling of surface reflectance, whereas the discarded ocean-colour, fluorescence, and oxygen- and water-vapour-absorption bands serve the sensor's marine and atmospheric mission and add correlated input dimensions without proportional burn-discriminative power.

\paragraph{Architecture comparison}
U-Net and SegFormer perform on par. Under the band-subset configuration, their temporal pre+post variants reach the same F1 ($68.24$ per bbox; $68.68$ vs $68.38$ per event, within one standard deviation), with U-Net obtaining the higher IoU ($59.33$ vs $58.72$); under the full-band configuration, SegFormer leads by $0.99$ F1 ($67.64$ vs $66.65$), within one standard deviation of the U-Net runs. SegFormer shows lower run-to-run variability in this setting ($\pm0.91$ and $\pm1.19$ F1, against $\pm1.68$ and $\pm1.72$ for U-Net). SegFormer is also the strongest spatial model, with the bi-temporal variant reaching $65.50$ and $65.73$ F1. ConvNeXt-UPerNet remains the weakest architecture, possibly because the cyclic repetition of ImageNet RGB filters is a rough adaptation to the multispectral OLCI input.

\paragraph{Qualitative analysis}
Figure~\ref{fig:qualitatives} shows per-timestep predictions of ConvLSTM + SegFormer with pre+post conditioning under both spectral configurations. \textbf{EMSR457 - AOI 1} (Figure~\ref{fig:emsr457}) is a successful case: over eight acquisitions, both models track a progressively growing fire. The full-band model is silent at $t_0$ and delineates the initial fire shape at $t_1$, whereas the band-subset model produces a small spurious detection at $t_0$ and misses the main extent at $t_1$; from $t_2$ onward both predictions grow consistently and converge to the ground-truth perimeter, remaining coherent even when partial cloud cover degrades the input at $t_5$. \textbf{EMSR454 - AOI 4} (Figure~\ref{fig:emsr454}) is a failure case: a small coastal burn with limited valid-pixel coverage, for which predictions are scattered across the three timesteps and never consistently overlap the scar, suggesting that a fire spanning only a handful of 300\,m pixels offers too little spectral contrast.

%% file: tables/all.tex
\begin{table*}[!ht]
\centering
\caption{Average F1 and IoU on the fire class for all model configurations, spectral settings, temporal conditioning modes, and backbone architectures, evaluated on the test set (mean $\pm$ standard deviation over three seeds; best value per column in bold). For 2D models, the pre- and post-fire input is the channel-wise concatenation of the pre-fire and the final post-fire acquisition (bi-temporal early fusion).}
\label{tab:all}
\resizebox{\textwidth}{!}{%
\begin{tabular}{lllcccc}
\toprule
\multirow{2}{*}{Model} & \multirow{2}{*}{Bands} & \multirow{2}{*}{Input} & \multicolumn{2}{c}{F1} & \multicolumn{2}{c}{IoU} \\
\cmidrule(lr){4-5} \cmidrule(lr){6-7}
 & & & 2D & 3D & 2D & 3D \\
\midrule
\multirow{4}{*}{ConvNeXt--UPerNet}
 & \multirow{2}{*}{5}  & Post-fire only     & $55.86 \pm 0.79$ & $58.82 \pm 1.02$ & $47.66 \pm 0.55$ & $49.69 \pm 0.94$ \\
 &                     & Pre- and post-fire & $58.71 \pm 1.65$ & $59.80 \pm 1.25$ & $50.26 \pm 0.92$ & $51.39 \pm 0.92$ \\
\cmidrule(lr){2-7}
 & \multirow{2}{*}{21} & Post-fire only     & $55.99 \pm 0.56$ & $56.92 \pm 1.30$ & $48.02 \pm 0.45$ & $48.01 \pm 1.00$ \\
 &                     & Pre- and post-fire & $57.26 \pm 1.81$ & $59.98 \pm 1.42$ & $49.17 \pm 1.43$ & $51.17 \pm 1.21$ \\
\midrule
\multirow{4}{*}{U-Net}
 & \multirow{2}{*}{5}  & Post-fire only     & $63.37 \pm 0.83$ & $62.73 \pm 1.08$ & $54.75 \pm 0.86$ & $53.74 \pm 0.79$ \\
 &                     & Pre- and post-fire & $64.64 \pm 1.11$ & $\mathbf{68.24 \pm 1.72}$ & $56.24 \pm 0.90$ & $\mathbf{59.33 \pm 1.52}$ \\
\cmidrule(lr){2-7}
 & \multirow{2}{*}{21} & Post-fire only     & $62.20 \pm 1.14$ & $62.63 \pm 2.70$ & $53.66 \pm 0.74$ & $53.45 \pm 2.67$ \\
 &                     & Pre- and post-fire & $64.63 \pm 2.31$ & $66.65 \pm 1.68$ & $56.00 \pm 2.08$ & $57.60 \pm 1.74$ \\
\midrule
\multirow{4}{*}{SegFormer}
 & \multirow{2}{*}{5}  & Post-fire only     & $63.64 \pm 0.26$ & $63.54 \pm 1.61$ & $55.22 \pm 0.31$ & $54.43 \pm 1.49$ \\
 &                     & Pre- and post-fire & $\mathbf{65.73 \pm 0.10}$ & $\mathbf{68.24 \pm 1.19}$ & $\mathbf{56.90 \pm 0.19}$ & $58.72 \pm 1.05$ \\
\cmidrule(lr){2-7}
 & \multirow{2}{*}{21} & Post-fire only     & $63.96 \pm 1.69$ & $63.91 \pm 1.76$ & $55.07 \pm 1.53$ & $54.63 \pm 1.53$ \\
 &                     & Pre- and post-fire & $65.50 \pm 0.83$ & $67.64 \pm 0.91$ & $56.73 \pm 1.03$ & $58.47 \pm 0.92$ \\
\bottomrule
\end{tabular}%
}
\end{table*}

%% file: sections/06_conclusions.tex
\section{Conclusions}
\label{sec:conclusions}
In this work, we presented TMB-S3, a large-scale burn scar delineation dataset comprising 246 European wildfire events with temporally paired Sentinel-3 OLCI acquisitions. We investigated the temporal aspect of burn scar segmentation on Sentinel-3 OLCI imagery, empirically finding that explicitly modelling time with a ConvLSTM encoder yields consistent gains on matched pre- and post-fire acquisitions. With post-fire-only sequences, the recurrent encoder brings no consistent benefit. A bi-temporal early-fusion baseline shows that the pre-fire reference is beneficial also for spatial models, while modelling the full post-fire sequence brings a further, consistent improvement. Limitations include possible spatial leakage, as nearby events from the same geographic cluster can fall in different splits, the European extent of the study, the lack of evaluation on fire-free scenes, and the absence of attention-based temporal encoders.


%% file: refs.bib
@inproceedings{xie2021segformer,
 author = {Xie, Enze and Wang, Wenhai and Yu, Zhiding and Anandkumar, Anima and Alvarez, Jose M. and Luo, Ping},
 booktitle = {Advances in Neural Information Processing Systems},
 editor = {M. Ranzato and A. Beygelzimer and Y. Dauphin and P.S. Liang and J. Wortman Vaughan},
 pages = {12077--12090},
 publisher = {Curran Associates, Inc.},
 title = {SegFormer: Simple and Efficient Design for Semantic Segmentation with Transformers},
 volume = {34},
 year = {2021}
}

@inproceedings{liu2022convnext,
  author    = {Liu, Zhuang and Mao, Hanzi and Wu, Chao-Yuan and Feichtenhofer, Christoph and Darrell, Trevor and Xie, Saining},
  title     = {A {ConvNet} for the 2020s},
  booktitle = {Proceedings of the IEEE/CVF Conference on Computer Vision and Pattern Recognition (CVPR)},
  pages     = {11976--11986},
  year      = {2022}
}

@dataset{zanaga2022worldcover,
  author    = {Zanaga, Daniele and Van De Kerchove, Ruben and Daems, Dirk and De Keersmaecker, Wanda and Brockmann, Carsten and Kirches, Grit and others},
  title     = {{ESA WorldCover} 10 m 2021 v200},
  year      = {2022},
  doi       = {10.5281/zenodo.7254221},
  publisher = {Zenodo}
}

@inproceedings{convlstm,
 author = {SHI, Xingjian and Chen, Zhourong and Wang, Hao and Yeung, Dit-Yan and Wong, Wai-kin and WOO, Wang-chun},
 booktitle = {Advances in Neural Information Processing Systems},
 editor = {C. Cortes and N. Lawrence and D. Lee and M. Sugiyama and R. Garnett},
 pages = {},
 publisher = {Curran Associates, Inc.},
 title = {Convolutional LSTM Network: A Machine Learning Approach for Precipitation Nowcasting},
 volume = {28},
 year = {2015}
}

@InProceedings{10.xiao2018upernet,
author="Xiao, Tete
and Liu, Yingcheng
and Zhou, Bolei
and Jiang, Yuning
and Sun, Jian",
editor="Ferrari, Vittorio
and Hebert, Martial
and Sminchisescu, Cristian
and Weiss, Yair",
title="Unified Perceptual Parsing for Scene Understanding",
booktitle="Computer Vision -- ECCV 2018",
year="2018",
publisher="Springer International Publishing",
address="Cham",
pages="432--448",
isbn="978-3-030-01228-1"
}

@incollection{key2006nbr,
  title     = {Landscape Assessment (LA): Ground measure of severity, the Composite Burn Index; and Remote sensing of severity, the Normalized Burn Ratio},
  author    = {Key, Carl H. and Benson, Nathan C.},
  booktitle = {FIREMON: Fire Effects Monitoring and Inventory System},
  series    = {Gen. Tech. Rep. RMRS-GTR-164-CD},
  publisher = {USDA Forest Service, Rocky Mountain Research Station},
  address   = {Ogden, UT},
  year      = {2006}
}

@article{giglio2018mcd64,
  title   = {The Collection 6 {MODIS} burned area mapping algorithm and product},
  author  = {Giglio, Louis and Boschetti, Luigi and Roy, David P. and Humber, Michael L. and Justice, Christopher O.},
  journal = {Remote Sensing of Environment},
  volume  = {217},
  pages   = {72--85},
  year    = {2018},
  doi     = {10.1016/j.rse.2018.08.005}
}

@article{lizundia2020firecci,
  title   = {A spatio-temporal active-fire clustering approach for global burned area mapping at 250\,m from {MODIS} data},
  author  = {Lizundia-Loiola, Joshua and Ot{\'o}n, Gonzalo and Ramo, Rub{\'e}n and Chuvieco, Emilio},
  journal = {Remote Sensing of Environment},
  volume  = {236},
  pages   = {111493},
  year    = {2020},
  doi     = {10.1016/j.rse.2019.111493}
}

@article{knopp2020unet,
  title   = {A Deep Learning Approach for Burned Area Segmentation with {Sentinel-2} Data},
  author  = {Knopp, Lisa and Wieland, Marc and R{\"a}ttich, Michaela and Martinis, Sandro},
  journal = {Remote Sensing},
  volume  = {12},
  number  = {15},
  pages   = {2422},
  year    = {2020},
  doi     = {10.3390/rs12152422}
}

@article{seydi2022burntnet,
  title   = {Burnt-Net: Wildfire burned area mapping with single post-fire {Sentinel-2} data and deep learning morphological neural network},
  author  = {Seydi, Seyd Teymoor and Hasanlou, Mahdi and Chanussot, Jocelyn},
  journal = {Ecological Indicators},
  volume  = {140},
  pages   = {108999},
  year    = {2022},
  doi     = {10.1016/j.ecolind.2022.108999}
}

@article{zhang2021synergy,
  title   = {Deep-learning-based burned area mapping using the synergy of {Sentinel-1} and {Sentinel-2} data},
  author  = {Zhang, Qi and Ge, Linlin and Zhang, Ruiheng and Metternicht, Graciela Isabel and Du, Zheyuan and Kuang, Jianming and Xu, Min},
  journal = {Remote Sensing of Environment},
  volume  = {264},
  pages   = {112575},
  year    = {2021},
  doi     = {10.1016/j.rse.2021.112575}
}

@article{sui2024biaunet,
  title   = {{BiAU-Net}: Wildfire burnt area mapping using bi-temporal {Sentinel-2} imagery and {U-Net} with attention mechanism},
  author  = {Sui, Tang and Huang, Qunying and Wu, Mingda and Wu, Meiliu and Zhang, Zhou},
  journal = {International Journal of Applied Earth Observation and Geoinformation},
  volume  = {132},
  pages   = {104034},
  year    = {2024},
  doi     = {10.1016/j.jag.2024.104034}
}

@article{xu2023slstr,
  title   = {Sentinel-3 {SLSTR} active fire detection and {FRP} daytime product -- Algorithm description and global intercomparison to {MODIS}, {VIIRS} and {Landsat} {AF} data},
  author  = {Xu, Weidong and Wooster, Martin J.},
  journal = {Science of Remote Sensing},
  volume  = {7},
  pages   = {100087},
  year    = {2023},
  doi     = {10.1016/j.srs.2023.100087}
}

@article{pinto2020banet,
  title   = {A deep learning approach for mapping and dating burned areas using temporal sequences of satellite images},
  author  = {Pinto, Miguel M. and Libonati, Renata and Trigo, Ricardo M. and Trigo, Isabel F. and DaCamara, Carlos C.},
  journal = {ISPRS Journal of Photogrammetry and Remote Sensing},
  volume  = {160},
  pages   = {260--274},
  year    = {2020},
  doi     = {10.1016/j.isprsjprs.2019.12.014}
}

@article{pinto2021practical,
  title   = {A Practical Method for High-Resolution Burned Area Monitoring Using {Sentinel-2} and {VIIRS}},
  author  = {Pinto, Miguel M. and Trigo, Ricardo M. and Trigo, Isabel F. and DaCamara, Carlos C.},
  journal = {Remote Sensing},
  volume  = {13},
  number  = {9},
  pages   = {1608},
  year    = {2021},
  doi     = {10.3390/rs13091608}
}

@article{sdraka2024floga,
  title   = {{FLOGA}: A Machine-Learning-Ready Dataset, a Benchmark, and a Novel Deep Learning Model for Burnt Area Mapping With {Sentinel-2}},
  author  = {Sdraka, Maria and Dimakos, Alkinoos and Malounis, Alexandros and Ntasiou, Zisoula and Karantzalos, Konstantinos and Michail, Dimitrios and Papoutsis, Ioannis},
  journal = {IEEE Journal of Selected Topics in Applied Earth Observations and Remote Sensing},
  volume  = {17},
  pages   = {7801--7824},
  year    = {2024},
  doi     = {10.1109/JSTARS.2024.3381737}
}

@article{cambrin2024cabuar,
  title   = {{CaBuAr}: California burned areas dataset for delineation},
  author  = {Rege Cambrin, Daniele and Colomba, Luca and Garza, Paolo},
  journal = {IEEE Geoscience and Remote Sensing Magazine},
  volume  = {11},
  number  = {3},
  pages   = {106--113},
  year    = {2023},
  doi     = {10.1109/MGRS.2023.3292467}
}

@inproceedings{colomba2022dataset,
  title     = {A Dataset for Burned Area Delineation and Severity Estimation from Satellite Imagery},
  author    = {Colomba, Luca and Farasin, Alessandro and Monaco, Simone and Greco, Salvatore and Garza, Paolo and Apiletti, Daniele and Baralis, Elena and Cerquitelli, Tania},
  booktitle = {Proceedings of the 31st ACM International Conference on Information and Knowledge Management (CIKM)},
  pages     = {3893--3897},
  year      = {2022},
  doi       = {10.1145/3511808.3557528}
}

@article{russwurm2018recurrent,
  title   = {Multi-Temporal Land Cover Classification with Sequential Recurrent Encoders},
  author  = {Ru{\ss}wurm, Marc and K{\"o}rner, Marco},
  journal = {ISPRS International Journal of Geo-Information},
  volume  = {7},
  number  = {4},
  pages   = {129},
  year    = {2018},
  doi     = {10.3390/ijgi7040129}
}

@inproceedings{garnot2020psetae,
  title     = {Satellite Image Time Series Classification with Pixel-Set Encoders and Temporal Self-Attention},
  author    = {Sainte Fare Garnot, Vivien and Landrieu, Loic and Giordano, Sebastien and Chehata, Nesrine},
  booktitle = {Proceedings of the IEEE/CVF Conference on Computer Vision and Pattern Recognition (CVPR)},
  pages     = {12325--12334},
  year      = {2020}
}

@inproceedings{garnot2021utae,
  title     = {Panoptic Segmentation of Satellite Image Time Series with Convolutional Temporal Attention Networks},
  author    = {Sainte Fare Garnot, Vivien and Landrieu, Loic},
  booktitle = {Proceedings of the IEEE/CVF International Conference on Computer Vision (ICCV)},
  pages     = {4872--4881},
  year      = {2021},
  doi       = {10.1109/ICCV48922.2021.00483}
}

@article{chen2021bit,
  title   = {Remote Sensing Image Change Detection with Transformers},
  author  = {Chen, Hao and Qi, Zipeng and Shi, Zhenwei},
  journal = {IEEE Transactions on Geoscience and Remote Sensing},
  volume  = {60},
  pages   = {1--14},
  year    = {2021},
  doi     = {10.1109/TGRS.2021.3095166}
}

@inproceedings{gerard2023wildfirespreadts,
  title     = {{WildfireSpreadTS}: A Dataset of Multi-Modal Time Series for Wildfire Spread Prediction},
  author    = {Gerard, Sebastian and Zhao, Yu and Sullivan, Josephine},
  booktitle = {Advances in Neural Information Processing Systems (NeurIPS), Datasets and Benchmarks Track},
  volume    = {36},
  year      = {2023}
}

@article{farasin2020double,
  title={Double-step u-net: A deep learning-based approach for the estimation of wildfire damage severity through sentinel-2 satellite data},
  author={Farasin, Alessandro and Colomba, Luca and Garza, Paolo},
  journal={Applied Sciences},
  volume={10},
  number={12},
  pages={4332},
  year={2020},
  publisher={MDPI}
}

@inproceedings{arnaudo2023robust,
  title={Robust burned area delineation through multitask learning},
  author={Arnaudo, Edoardo and Barco, Luca and Merlo, Matteo and Rossi, Claudio},
  booktitle={Joint European Conference on Machine Learning and Knowledge Discovery in Databases},
  pages={436--447},
  year={2023},
  organization={Springer}
}
